\documentclass[
]{ceurart}

\usepackage{listings}
\usepackage{multirow}
\usepackage{enumitem}
\usepackage{graphicx}
\usepackage{tabularx}
\usepackage{subfig}
\usepackage{booktabs, multirow} 
\begin{document}

\copyrightyear{2023}
\copyrightclause{Copyright for this paper by its authors.
  Use permitted under Creative Commons License Attribution 4.0
  International (CC BY 4.0).}

\conference{Forum for Information Retrieval Evaluation, December 15-18, 2023, India}


\title{Harnessing Pre-Trained Sentence Transformers for Offensive Language Detection in Indian Languages}

\author[1, 3]{Ananya Joshi}[%
email=joshiananya20@gmail.com
]

\author[2, 3]{Raviraj Joshi}[%
email=ravirajoshi@gmail.com
]

\address[1]{MKSSS Cummins College of Engineering for Women, Pune, Maharashtra, India}
\address[2]{Indian Institute of Technology Madras, Chennai, Tamil Nadu, India}
\address[3]{L3Cube Pune, India}

\begin{abstract}
  In our increasingly interconnected digital world, social media platforms have emerged as powerful channels for the dissemination of hate speech and offensive content. This work delves into the domain of hate speech detection, placing specific emphasis on three low-resource Indian languages: Bengali, Assamese, and Gujarati. The challenge is framed as a text classification task, aimed at discerning whether a tweet contains offensive or non-offensive content. Leveraging the HASOC 2023 datasets, we fine-tuned pre-trained BERT and SBERT models to evaluate their effectiveness in identifying hate speech. Our findings underscore the superiority of monolingual sentence-BERT models, particularly in the Bengali language, where we achieved the highest ranking. However, the performance in Assamese and Gujarati languages signifies ongoing opportunities for enhancement. Our goal is to foster inclusive online spaces by countering hate speech proliferation.
\end{abstract}


\maketitle

\section{Introduction}
In today's interconnected world, social media platforms have gained significant influence and have become powerful means of spreading hate speech, often targeting individuals or groups based on factors like race, caste, gender, sexual orientation, or political beliefs. The negative effects of this trend, including cyberbullying and the presence of offensive content, are well-documented and can harm the mental well-being of users. As the number of people using social media continues to grow, it's crucial to develop effective methods to identify and address offensive language to maintain a positive online environment.

Efficient tools for detecting offensive, vulgar, and hateful language on social media platforms are essential because such language can disrupt online discussions and have real-world consequences. This highlights the need for robust Natural Language Processing (NLP) systems capable of effectively recognizing and countering offensive language on these platforms \cite{velankar2022review}.

Our research specifically focuses on the challenge of detecting offensive, profane, and hateful language in low-resource Indian languages, namely Assamese, Bengali, and Gujarati. These languages have received relatively less attention in the field of NLP, and they each have unique linguistic characteristics that require specialized solutions for addressing offensive content effectively.

Bengali, primarily spoken in West Bengal, India, and Bangladesh, is known for its rich literary tradition and cultural significance. With over 230 million speakers, it ranks as the second most spoken language in India and the seventh in the world. Gujarati, predominantly spoken in the Indian state of Gujarat, contributes significantly to India's linguistic diversity with approximately 55 million speakers. Assamese, spoken primarily in the northeastern Indian state of Assam, is rooted in Sanskrit and includes various dialects, playing a vital role in the linguistic diversity of India's northeastern region.
\\\\
Hate Speech and Offensive Content Identification in English and Indo-Aryan Languages (HASOC) 2023\footnote[1]{\url{https://hasocfire.github.io/hasoc/2023/}} initiative includes four distinct tasks. \\We specifically concentrate on two tasks:
\begin{itemize}
\item[$\bullet$] Task 1B: Identifying Hate, Offensive, and Profane Content in Gujarati
\item[$\bullet$] Task 4: Annihilate Hates - Detecting Hate Speech in Bengali \cite{annihilate-hates-bengali} and Assamese \cite{annihilate-hates-assamese}
\end{itemize}

Throughout this paper, we rigorously evaluate the performance of both single-language and multi-language models when applied to the datasets associated with these tasks. We primarily focus on sentence-BERT models for identifying offensive language in social media contexts, showcasing their superior performance. Notably, we present state-of-the-art results on the HASOC 2023 test set using specialized models such as BengaliSBERT, GujaratiSBERT \cite{deode2023l3cubeindicsbert}, and assamese-bert \cite{joshi2022l3cube_hindbert}, which have been developed by L3Cube-Pune\footnote[2]{\url{https://huggingface.co/l3cube-pune}}.

\section{Related Work}
The task of hate speech detection in multilingual contexts has garnered significant attention in recent shared tasks and research endeavors. In this section, we provide an overview of related work, with a focus on studies relevant to our investigation of hate speech detection in low-resource Indian languages.

Several shared tasks have aimed to address hate speech detection challenges. For instance, the paper \cite{chakravarthi2021developing} analyses the systems submitted for the HASOC shared tasks and DravidianLangTech workshop conducted in 2020, focusing on Malayalam, Tamil, and Kannada offensive posts on social media. \cite{ranasinghe2022overview} describes the Subtrack 3 of HASOC-2022, focusing on Offensive Language Identification in Marathi. \cite{satapara2021overview} describes the HASOC 2021 subtask of identification of conversational hate speech in code-mixed languages.

Hindi and Marathi, two prominent Indian languages, have received considerable attention in hate speech detection research. Notable studies include \cite{velankar2021hate, chavan2022twitter, velankar2022l3cube, Ghosh2022BaselineBM, 10.1145/3576913}, which have contributed to the understanding of hate speech dynamics in these languages. \cite{velankar2022mono} presents a comparative study between monolingual and multilingual BERT models for hate speech detection in Marathi language, while \cite{ghosh-senapati-2022-hate} presents a similar comparative analysis with cross-language evaluation for Hindi and Marathi.

While Hindi and Marathi have been extensively studied, research efforts have expanded to include languages such as Bengali and Assamese. \cite{das2022hate} offers insights into hate speech detection in Bengali, while \cite{10183497} presents transformer based hate speech detection in Assamese. Similar challenges have been explored in South Indian languages, adding to the linguistic diversity of hate speech research. \cite{ROY2022101386} suggests a weighted ensemble framework to capture hate speech and offensive languages on social platforms posted in code-mixed languages like Hindi–English, Tamil–English, Malayalam–English, Telugu–English, and others. The paper \cite{sai2021towards} proposes a novel technique of selective translation and transliteration for code-mixed and romanized offensive speech classification in Dravidian languages. 

These prior studies provide valuable foundations for our investigation into hate speech detection in low-resource Indian languages, such as Assamese, Bengali, and Gujarati, underscoring the growing recognition of the need to address hate speech in diverse linguistic contexts.

\section{Experimental Setup}

\subsection{Task description}

Below, we provide an overview of the tasks:
\begin{itemize}
\item[$\bullet$] \textbf{Task 1B: Identifying Hate, offensive and profane content in Gujarati}\footnote[3]{\url{https://hasocfire.github.io/hasoc/2023/task1.html}}: \\
This task focuses on Hate speech and Offensive language identification for Gujarati. This is a coarse-grained binary classification in a few-shot setting, in which participating systems are required to classify tweets into two classes, namely: Hate and Offensive (HOF) and Non-Hate and offensive (NOT).

\begin{itemize}
\item[>]\textbf{(NOT) Non Hate-Offensive} - This post does not contain any hate speech, profane, offensive content. 
\item[>]\textbf{(HOF) Hate and Offensive} - This post contains hate, offensive, and profane content.
\end{itemize}

\item[$\bullet$] \textbf{Task 4: Annihilate Hates\footnote[4]{\url{https://sites.google.com/view/hasoc-2023-annihilate-hates/home}}}:\\
The objective of the task is to detect hate speech in Bengali, Bodo, and Assamese languages. It is a binary classification task. Each dataset (for the three languages) consists of a list of sentences with their corresponding class (hate or offensive (HOF) or not hate (NOT)). Data is primarily collected from Twitter, Facebook, or youtube comments. Team rank is determined based on the Macro F1 score.
\end{itemize}

\subsection{Datasets}

\begin{figure}[]
\centering
\includegraphics[scale=0.5]{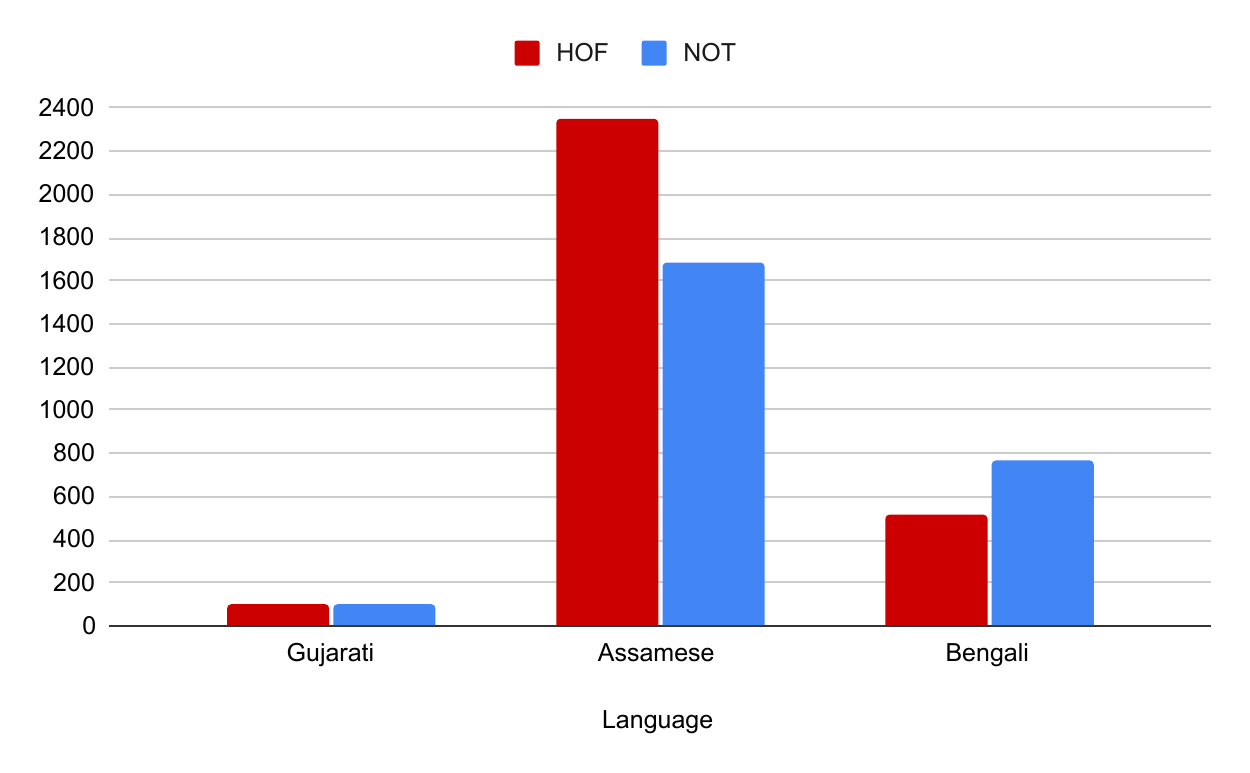}
\caption{Class distribution of tweets in HASOC 2023 training datasets for Task 1 and Task 4}\label{fig:1}
\end{figure}

HASOC 2023 provides training datasets tagged as "NOT" and "HOF" for binary classification for both Task 1 and Task 4. The main source of data collection is Twitter, Facebook, or YouTube comments. Table \ref{tab:1} shows all dataset statistics. The distribution of offensive and non-offensive tweets in the training dataset of each language is depicted in Figure \ref{fig:1}

\begin{table}[!htp]\centering
\caption{HASOC 2023 datasets statistics for Task 1 and Task 4}\label{tab:1}
\scriptsize
\begin{tabular}{|l|c|ccc|c|}\toprule
&\textbf{} &\multicolumn{3}{c|}{\textbf{Training}} &\textbf{Test} \\\midrule
&Label -> &HOF &NOT &Total &Total \\ \midrule
Task 1 &Gujarati &100 &100 &200 &1196 \\ \midrule
\multirow{2}{*}{Task 4} &Assamese &2347 &1689 &4036 &1009 \\
&Bengali &515 &766 &1281 &320 \\
\bottomrule
\end{tabular}
\end{table}

\subsection{Preprocessing}
In order to enhance the accuracy of our classification task, we conducted data preprocessing to improve the data quality. We engaged in cleaning procedures to optimize the data conditions, which included eliminating punctuation marks, URLs, usernames, handles, hashtags, numbers, and Roman characters. Additionally, our preprocessing methods addressed issues such as newline characters, excessive spaces, and empty parentheses. Notably, we made a deliberate decision to retain emojis, as they contribute significantly to conveying the sentiment of the text and were observed to yield superior results.
\\\\
Label encoding: We encode Class label into a unique number for each task: "HOF" to "1", and "NOT" to "0"

\subsection{Models and Training Setup}

BERT \cite{devlin2019bert} models are pre-trained on a massive corpus of text data, where they learn to predict masked words within sentences. Then, they are fine-tuned on specific downstream tasks using labeled data.
Sentence-BERT (SBERT) \cite{reimers2019sentence} models are trained by learning fixed-size embeddings for sentences using siamese or triplet network architectures that aim to optimize similarity scores between related sentences and minimize distances between them in embedding space. 

While BERT focuses on word-level representations, SBERT models are designed to capture the semantic meaning of sentences, including subtle nuances and context, by producing fixed-size sentence embeddings. Hate speech often relies on the overall context and phrasing of a sentence, making SBERT's sentence-level understanding more relevant. Hate speech classification often requires an understanding of context and context-dependent variations in meaning. SBERT models, leverage contextual information by considering the surrounding words in a sentence, making them more adept at recognizing the intended sentiment or tone. Traditional BERT models, while powerful, may struggle to understand the nuances of entire sentences and their emotional or hateful intent.

The papers \cite{joshi2022l3cubemahasbert, deode2023l3cubeindicsbert} show that the Sentence-BERT models outperform the corresponding BERT variants in understanding context-specific information. Hence, we primarily utilize the monolingual and multilingual SBERT models for Gujarati and Bengali languages. The Assamese language, however, lacks quality datasets and powerful models such as Sentence-BERT. Hence, we use the monolingual assamese-bert and the multilingual indic-bert model.

\begin{itemize}
\item[$\bullet$] For Task 1, we use the pre-trained monolingual model GujaratiSBERT\footnote[5]{\url{https://huggingface.co/l3cube-pune/gujarati-sentence-bert-nli}} and the multilingual IndicSBERT\footnote[6]{\url{https://huggingface.co/l3cube-pune/indic-sentence-bert-nli}} model. 
\item[$\bullet$] For Task 4, we use pre-trained monolingual models of BengaliSBERT\footnote[7]{\url{https://huggingface.co/l3cube-pune/bengali-sentence-bert-nli}}, bengali-bert\footnote[8]{\url{https://huggingface.co/l3cube-pune/bengali-bert}}, assamese-bert\footnote[9]{\url{https://huggingface.co/l3cube-pune/assamese-bert}} and the multilingual models IndicSBERT, indic-bert\footnote[10]{\url{https://huggingface.co/ai4bharat/indic-bert}}.
\end{itemize}

For both tasks, we initialize a classification model using the BERT architecture and freeze the first six layers of the model. Next, we train the model using the provided training data for 4 epochs with the default learning rate.

\section{Results}
We conducted training on a range of models using the complete training dataset and subsequently employed these models to predict classes for the provided test dataset.  In all the tasks, the texts are classified into 2 categories- HOF, indicating the presence of hateful content, or NOT- indicating no offensive content. The outcomes are presented in Table \ref{tab:2}, and the evaluation metric employed for determining the team's leaderboard ranking was the Macro F1 Score. We have included all the task results in accordance with the leaderboard presentation. Additionally, we explored the efficacy of multiple pre-trained BERT and SBERT models but submitted only the most successful run for evaluation, omitting the submission of other runs due to their subpar performance.

We achieved the \textbf{top ranking (rank 1)} among 21 participating teams for Task 4- Bengali language, because of the highest Macro F1 Score obtained using the BengaliSBERT model. The BengaliSBERT model outperforms the bengali-bert and other multilingual models like MuRil, Indic-bert and IndicSBERT. 
For Task 4- Assamese language, we attained rank 6 among 20 teams through the use of assamese-bert model. For Task1- Gujarati, we stand at Rank 10 among 17 participating teams. The best score was given by GujaratiSBERT model, outperforming the gujarati-bert and other multilingual models like MuRil, Indic-bert and IndicSBERT.


\begin{table}[!htp]\centering
\caption{Macro F1 scores obtained from various models, along with the Ranks achieved in Task1 and Task4 of HASOC 2023}\label{tab:2}
\scriptsize
\begin{tabular}{|ccccc|}\toprule
\textbf{Task} &\textbf{Language} &\textbf{Model} &\textbf{MACRO F1} &\textbf{Rank} \\\midrule
\multirow{2}{*}{Task 1} &\multirow{2}{*}{Gujarati} &GujaratiSBERT &0.7324 &10 \\
& &IndicSBERT &0.7291 & \\ \midrule
\multirow{2}{*}{Task 4} &\multirow{2}{*}{Assamese} &assamese-bert &0.7065 &6 \\
& &indic-bert &0.6788 & \\ \midrule
\multirow{3}{*}{Task 4} &\multirow{3}{*}{Bengali} &BengaliSBERT &0.7703 &1 \\
& &IndicSBERT &0.7409 & \\ 
& &indic-bert &0.7121 & \\
\bottomrule
\end{tabular}
\end{table}

\section{Conclusion}
Through this paper, we describe our approach for hate and offensive speech detection in three Indian languages. We utilize the HASOC 2023 datasets for fine-tuning the pretrained BERT and SBERT models and testing their performance. Our findings reveal that monolingual Sentence-BERT models consistently outperform both monolingual BERT models and multilingual counterparts in the realm of hate speech identification. Notably, we secured the highest ranking for Bengali language, while the lower rankings in Assamese and Gujarati languages underscore the ongoing need for enhancements in these domains. Looking ahead, we are committed to exploring various strategies to elevate the performance of Assamese and Gujarati models. Our overarching goal is to contribute to the advancement of more inclusive and comprehensive tools for combatting online hate speech, ultimately fostering online spaces characterized by tolerance and respect.



\begin{acknowledgments}
  This work was done under the L3Cube Pune mentorship
program. We would like to express our gratitude towards
our mentors at L3Cube for their continuous support and
encouragement. 
\end{acknowledgments}

\bibliography{main}

\end{document}